\documentclass[11pt]{article}

\usepackage[margin=1in]{geometry}
\usepackage{graphicx}
\usepackage{amsmath, amssymb}
\usepackage{hyperref}
\usepackage{caption}
\usepackage{float}
\usepackage{booktabs}
\usepackage{enumitem}
\usepackage[backend=biber, style=apa]{biblatex}
\begin{document}

\begin{center}
    \LARGE\bfseries
    Segmented Confidence Sequences and Multi-Scale Adaptive Confidence Segments for Anomaly Detection in Nonstationary Time Series
    \vspace{2ex}

    \large
    Muyan Anna Li, Aditi Gautam \\
    NVIDIA, DGXC Applied AI Lab \\
    Santa Clara, CA, USA \\
    \texttt{annali@nvidia.com}, \texttt{adgautam@nvidia.com} \\ 
    August 5th, 2025
\end{center}


\begin{abstract}
As time series data become increasingly prevalent in domains such as manufacturing, IT, and infrastructure monitoring, anomaly detection must adapt to nonstationary environments where statistical properties shift over time. Traditional static thresholds are easily rendered obsolete by regime shifts, concept drift, or multi-scale changes. To address these challenges, we introduce and empirically evaluate two novel adaptive thresholding frameworks: Segmented Confidence Sequences (SCS) and Multi-Scale Adaptive Confidence Segments (MACS). Both leverage statistical online learning and segmentation principles for local, contextually sensitive adaptation, maintaining guarantees on false alarm rates even under evolving distributions. Our experiments across Wafer Manufacturing benchmark datasets show significant F1-score improvement compared to traditional percentile and rolling quantile approaches. This work demonstrates that robust, statistically principled adaptive thresholds enable reliable, interpretable, and timely detection of diverse real-world anomalies.
\end{abstract}

\section{Introduction}
Time series data are ubiquitous across modern applications, from industrial process monitoring and predictive maintenance to financial markets and sensor-driven systems. Detecting anomalies—unusual patterns or behaviors that deviate from expected trends—is crucial for preventing faults, reducing risk, and ensuring operational reliability \parencite{chandola2009}. Unlike static datasets, time series often exhibit evolving behavior, including trends, seasonality, and abrupt regime shifts, making anomaly detection a particularly challenging problem.

In recent years, researchers have developed advanced techniques that go beyond simple static thresholds. Approaches such as robust moving windows, online quantile estimation, and confidence sequence theory have emerged to provide more adaptive and statistically principled anomaly detection \parencite{howard2021, wang2023}. These methods aim to balance computational efficiency with real-time adaptability, enabling detection systems to respond to changing data dynamics.

However, existing adaptive thresholding methods often struggle when data exhibit multiple temporal scales or sudden regime shifts. Fixed-window or global percentile-based strategies may either fail to capture local variations, leading to missed anomalies, or produce excessive false positives when the baseline drifts \parencite{benidis2022}. This highlights the need for a thresholding framework that can simultaneously adapt to both abrupt and gradual changes in data distribution.

To address these challenges, we contribute two novel frameworks for adaptive thresholding:
\begin{itemize}[itemsep=0pt, parsep=0pt]
\item Segmented Confidence Sequences (SCS) segments time series by regime, maintaining distinct confidence-based bounds per segment, and adapts to local rather than global statistics.
\item Multi-Scale Adaptive Confidence Segments (MACS) is an approach that adapts detection simultaneously at multiple window lengths, enabling the detection of both rapid bursts and slow regime changes.
\item Comprehensive experimental evaluation supporting statistically significant improvements over traditional percentile or fixed adaptive thresholds.
\end{itemize}

\section{Related Work}

\subsection{Static and Traditional Thresholding}
Early approaches relied on fixed global thresholds – often prescribed as $\text{mean} \pm k\sigma$ or a static quantile – assuming stationarity and i.i.d. observations \parencite{chandola2009}. Although easy to implement, these methods fail under concept drift or dynamic variance and are prone to false positives in practical systems \parencite{blazquezgarcia2021}.

Percentile-based approaches, such as the 99th percentile threshold, adjust for heavy tails but still falter under persistent distributional drift or nonstationarity, as shown in benchmark studies \parencite{genton2021}. Methods based on Extreme Value Theory (EVT) and the Peak-Over-Threshold (POT) model the empirical tail beyond a high threshold but still assume the threshold regime is quasi-stationary \parencite{genton2021}.

\subsection{Sliding Windows, Rolling Statistics, and Moving Quantiles}
Adaptive methods using sliding windows recalculate thresholds over a recent window – updating the mean, standard deviation, or quantile in an online manner \parencite{aggarwal2015}. The exponential weighted moving average (EWMA) improves rapid adaptation to trends or regime switches, but window size determines sensitivity and is often hard to tune \parencite{blazquezgarcia2021}. Non-parametric dynamic models further reduce reliance on distributional assumptions and are superior in recall \parencite{rousseeuw2020}.

\subsection{Model-Based and Machine Learning Approaches}
Forecasting-model-based detection fits models such as ARIMA or seasonal decomposition, then tests for outliers in the model residuals \parencite{brockwell2016}. More advanced approaches leverage autoencoders, deep neural networks, or reinforcement learning agents to learn context-sensitive anomaly scores or directly optimize detection performance \parencite{benidis2022, ahmad2017, xue2023}. However, these methods either lack explicit statistical error guarantees or require considerable labeled anomaly data.

\subsection{Confidence Sequences for Online Adaptation}
Confidence sequences (CS) – time-uniform intervals guaranteeing correct coverage at all times – are a foundation for rigorous thresholding in nonstationary data, allowing error rate control under arbitrary stopping \parencite{howard2021}. RRecent algorithms can maintain confidence bounds for quantiles or means, enabling adaptive anomaly scoring robust to drift, heavy tails, or outliers \parencite{wang2023}. Applying CS-based threshold selection to streaming anomaly detection is a promising and newly emerging direction \parencite{howard2021, sun2024}.

\subsection{Segmentation-Based Local Thresholding}
Segmenting time series into locally stationary regimes – via APCA or clustering – brings statistical homogeneity to threshold estimation, allowing each regime to have a locally fitted, adaptive rule \parencite{keogh2001, aghabozorgi2015}. Recent approaches use clustering (e.g., k-means) on summary features to capture regime change, but statistical decision boundaries within each segment remain underexplored.

\section{Methods}

We focus on two novel, unsupervised adaptive thresholding strategies for streaming time series: Segmented Confidence Sequences (SCS) and Multi-Scale Adaptive Confidence Segments (MACS). Both are designed for practical anomaly detection pipelines (see Figure 1 and Figure 2).

\subsection{Segmented Confidence Sequences (SCS)}

SCS first performs time series segmentation using either Adaptive Piecewise Constant Approximation (APCA) - which iteratively splits at points that minimize reconstruction error - or feature-based K-means clustering using sliding-window statistics \parencite{aghabozorgi2015}. Each segment is assumed to be locally stationary, allowing for regime-specific anomaly detection. Within each segment, an independent confidence sequence is maintained for anomaly score thresholds, using Hoeffding's inequality for non-parametric bounds \parencite{howard2021}. Segment-specific anomaly flags are triggered if new scores exceed the upper confidence bound or fall below the lower confidence bound.

\begin{figure}[H]
    \centering
    \includegraphics[width=0.7\linewidth, height=1\textheight, keepaspectratio]{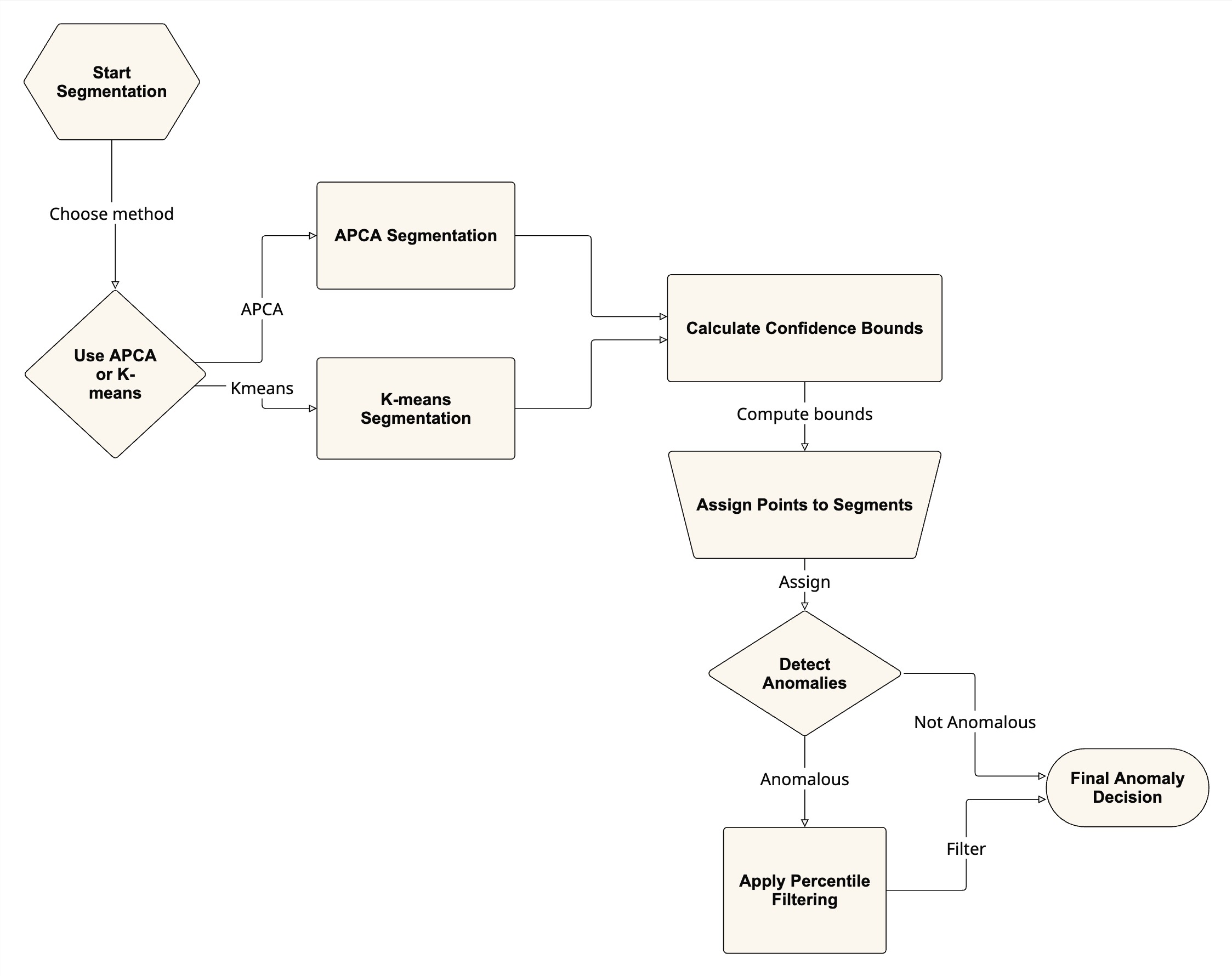}
    \caption{Illustration of the SCS flow}
    \label{fig:scs_flow}
\end{figure}

SCS begins by partitioning the time series into locally stationary segments, using either Adaptive Piecewise Constant Approximation (APCA) or feature-based K-means clustering \parencite{aghabozorgi2015}. APCA operates by iteratively identifying optimal split points that minimize total reconstruction error, defined as the sum of squared deviations from the mean within each segment. Specifically, for a proposed split, the reconstruction error is calculated as:

\[
\texttt{total\_error} = \texttt{left\_error} + \texttt{right\_error}
\]
\[
\texttt{left\_error} = \sum(x_i - \bar{x}_{\text{left}})^2 \quad\text{and}\quad \texttt{right\_error} = \sum(x_j - \bar{x}_{\text{right}})^2
\]

This process continues recursively until segments fall below a minimum length constraint or no further improvement is observed according to a specified threshold. For flat regions of the time series, identified by a coefficient of variation below 0.1, APCA defaults to fixed-length segmentation, where segment size is set to:

\[
\max(200, \left\lfloor \frac{n}{15} \right\rfloor)
\]

For more variable data, a candidate split is accepted only if the minimized reconstruction error satisfies:

\[
\texttt{min\_error} < \texttt{no\_split\_error} \times \texttt{improvement\_threshold}
\]

The improvement threshold is set to 0.7 for high-variance series and 0.5 for moderate-variance series.

Alternatively, SCS supports a K-means segmentation approach that clusters sliding window representations of the time series based on statistical features. For each window, features including the mean, standard deviation, median, and skewness are extracted, and the resulting feature vectors are normalized using \texttt{StandardScaler}. For multi-dimensional time series data, the dimensionality is reduced by averaging across the feature dimensions such that:

\[
\texttt{data\_1d} = \texttt{mean}(X, \texttt{axis}=1) \quad \text{if } X \in \mathbb{R}^{n \times d},\, d > 1
\]

In cases where the clustering process fails due to insufficient variability or degenerate distributions, the entire sequence is treated as a single segment to preserve stability.

Within each resulting segment, regardless of the segmentation method, SCS maintains an independent confidence sequence for thresholding anomaly scores. These bounds are derived using Hoeffding-style inequalities \parencite{howard2021} and are parameterized by the local standard deviation of the segment’s scores. The width of the confidence bound is initially set as:

\[
\texttt{bound\_width} = 1.5 \times \texttt{std\_score}
\]

It is then scaled by a factor that reflects the desired confidence level. Specifically, if the confidence level exceeds 95\%, the bound is widened by a factor of 1.2; if it is below 90\%, the bound is narrowed to 0.8. The final confidence interval for each score is given by:

\[
\texttt{lower\_bound} = \bar{x} - \texttt{bound\_width}, \quad \texttt{upper\_bound} = \bar{x} + \texttt{bound\_width}
\]

To ensure robustness and avoid false positives from local fluctuations, SCS uses a composite detection criterion: a point is flagged as anomalous only if it violates both the confidence bounds and a global percentile threshold. Formally, an intermediate anomaly indicator is computed as:

\[
\texttt{anomalies} = (x < \texttt{lower\_bound}) \vee (x > \texttt{upper\_bound})
\]

The final anomaly decision is made via:

\[
\texttt{final\_anomalies} = \texttt{anomalies} \wedge \texttt{percentile\_filter}
\]

To summarize, the algorithm flow is outlined below:

\begin{itemize}[itemsep=0pt, parsep=0pt]
    \item \textbf{Segmentation Phase}: Apply APCA or K-means to identify regime boundaries
    \item \textbf{Bound Calculation}: Compute confidence bounds for each segment independently
    \item \textbf{Point Assignment}: Dynamically assign incoming points to their corresponding segment
    \item \textbf{Anomaly Detection}: Compare each point to segment-specific thresholds
    \item \textbf{Filtering}: Apply percentile-based filtering for conservativeness
\end{itemize}

(The pseudocode of the algorithm flow is in \textit{Appendix A}.)

Incoming data points are dynamically assigned to their corresponding segment, and anomalies are detected by comparing each point to the segment-specific, adaptively updated threshold. This approach ensures that anomaly detection is locally calibrated to the current regime, providing robust detection even as the data distribution shifts over time. The method is unsupervised, requires no labeled anomalies, and is suitable for both batch and streaming data.

\subsection{Multi-Scale Adaptive Confidence Segments (MACS)}

MACS is designed to capture anomalies occurring at different temporal resolutions by maintaining multiple rolling windows of varying lengths in parallel. 

\begin{figure}[H]
    \centering
    \includegraphics[width=0.7\linewidth, height=0.6\textheight, keepaspectratio]{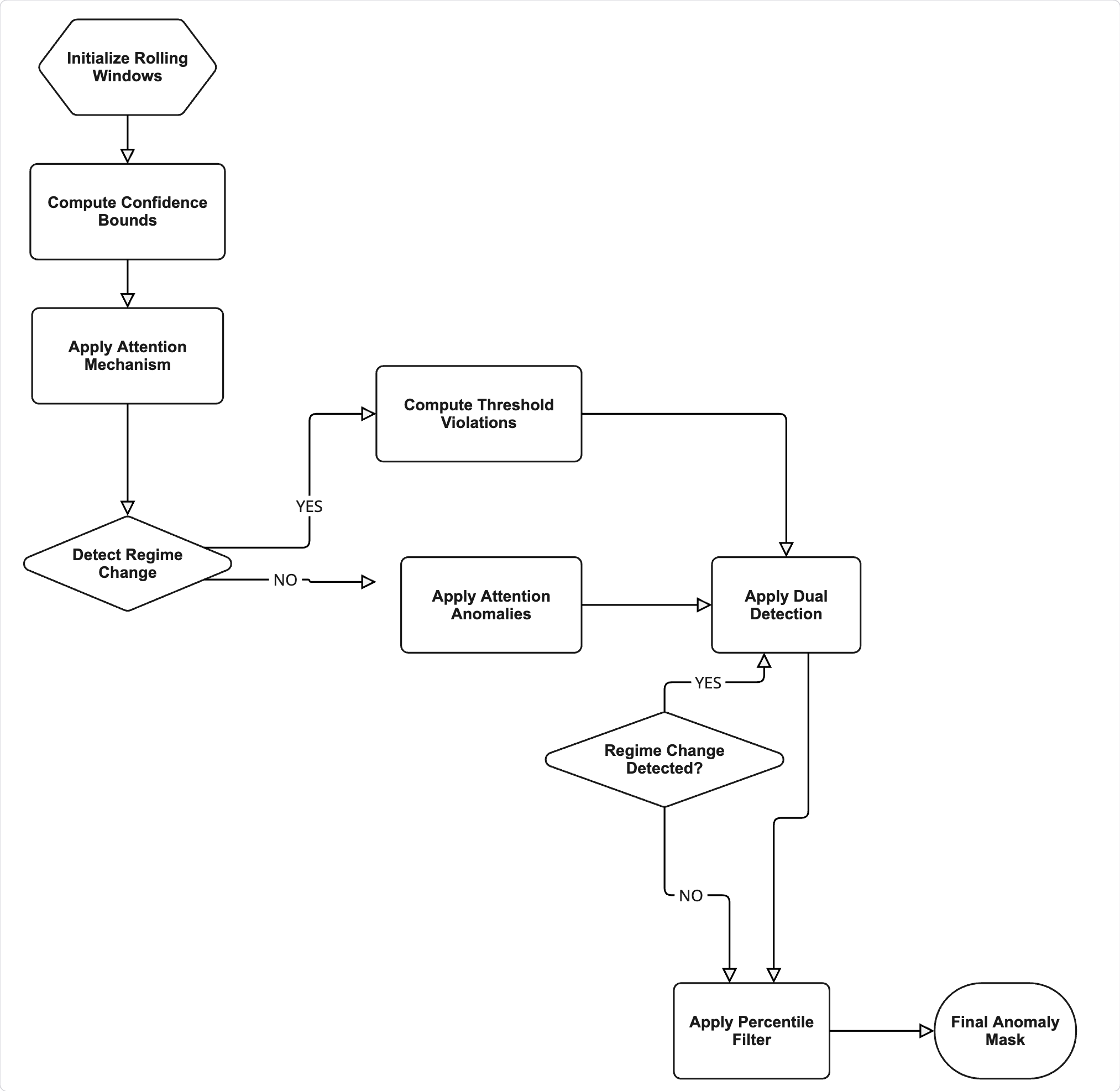}
    \caption{Illustration of the MACS flow}
    \label{fig:macs_flow}
\end{figure}

Specifically, it tracks short (e.g., 50 steps), medium (e.g., 100 steps), and long (e.g., 500 steps) time scales, each of which independently maintains a confidence sequence \parencite{blazquezgarcia2021}. This structure enables MACS to detect a broad spectrum of anomalies, from short-term bursts to slow-moving regime shifts. To further enhance adaptability, MACS incorporates an attention mechanism that dynamically weighs the importance of each temporal scale based on local variance patterns in the data.

Each temporal scale maintains its own confidence bounds, computed using the segment’s local statistics. For a given window, the width of the confidence bound is initialized as:

\[
\texttt{bound\_width} = 1.5 \times \texttt{std\_score}
\]

It is then scaled according to the desired confidence level. Specifically, the bound width is increased by 20\% for high-confidence settings (>95\%) and decreased by 20\% for low-confidence settings (<90\%). The final upper and lower bounds at each scale are then computed as:

\[
\texttt{lower\_bound} = \bar{x} - \texttt{bound\_width}, \quad \texttt{upper\_bound} = \bar{x} + \texttt{bound\_width}
\]

To integrate these multiple scales, MACS uses an attention mechanism that adjusts the relative importance of each scale based on the local variance of the scores. Local variance is estimated using a rolling variance window, defined as:

\[
\texttt{local\_variance} = \texttt{rolling\_var}(x, \texttt{window} = \min(\texttt{short\_window}, \left\lfloor n/10 \right\rfloor))
\]

Based on the level of local variance, different attention weights are assigned:
\begin{itemize}[itemsep=0pt, parsep=0pt]
    \item High variance ($> 0.7$): [0.6, 0.3, 0.1]
    \item Medium variance ($> 0.3$): [0.2, 0.6, 0.2]
    \item Low variance ($\leq 0.3$): [0.1, 0.3, 0.6]
\end{itemize}

These weights are used to compute a combined confidence bound as a weighted sum across scales:

\[
\texttt{combined\_bound} = \sum_{i=1}^{3} \texttt{weight}_i \cdot \texttt{bound}_i
\]

In addition to confidence sequences, MACS performs regime change detection using a CUSUM-like procedure based on rolling statistics. It tracks both the rolling mean and standard deviation over the long window. A regime change is flagged if the normalized change in mean exceeds 2.0, or if the change in standard deviation exceeds 1.5, defined respectively as:

\[
\texttt{mean\_change} = \frac{\bar{x}_{\text{current}} - \bar{x}_{\text{historical}}}{\texttt{std}_{\text{historical}} + 10^{-8}}, \quad
\texttt{std\_change} = \frac{\texttt{std}_{\text{current}} - \texttt{std}_{\text{historical}}}{\texttt{std}_{\text{historical}} + 10^{-8}}
\]

When a regime change is detected, MACS applies a conservative thresholding policy that requires agreement between two independent detection mechanisms.

The dual detection approach in MACS enhances robustness by combining two complementary strategies. First, a threshold violation counting mechanism flags a point as anomalous if it exceeds at least two out of three individual scale-specific thresholds:

\[
\texttt{threshold\_violations} = \sum_{i=1}^{3} \texttt{scale\_anomalies}_i \geq 2
\]

Second, MACS uses the attention-weighted combined bounds to detect deviations from the contextually prioritized envelope. A point is flagged as anomalous if its score lies outside this combined range:

\[
\texttt{attention\_anomalies} = (x < \texttt{combined\_lower}) \vee (x > \texttt{combined\_upper})
\]

The final decision rule is regime-aware. Under normal operating conditions, anomalies are flagged solely based on the attention-weighted bounds. However, during regime changes, both the threshold violation and the attention anomaly conditions must be satisfied simultaneously. Finally, MACS applies an additional percentile-based filter to avoid over-detection. This step discards low-magnitude outliers by requiring anomaly scores to exceed a global percentile threshold. The final anomaly mask is obtained as:

\[
\texttt{final\_anomalies} = \texttt{anomalies} \wedge \texttt{percentile\_filter}
\]

This layered structure – combining multi-scale bounds, adaptive attention, regime awareness, and statistical filtering – enables MACS to balance sensitivity and precision in diverse streaming environments effectively.

To summarize, the algorithm flow is outlined below:

\begin{itemize}[itemsep=0pt, parsep=0pt]
    \item \textbf{Multi-Scale Analysis}: Calculate confidence bounds at three temporal scales
    \item \textbf{Attention Calculation}: Compute local variance and determine attention weights
    \item \textbf{Bound Combination}: Apply attention mechanism to combine multi-scale bounds
    \item \textbf{Regime Detection}: Identify statistical regime changes using CUSUM-like logic
    \item \textbf{Dual Detection}: Apply both threshold violation counting and attention-weighted bounds
    \item \textbf{Regime-Aware Decision}: Combine detection methods based on regime state
    \item \textbf{Filtering}: Apply percentile-based filtering for conservativeness
\end{itemize}

(The pseudocode of the algorithm flow is in \textit{Appendix B}.)

\subsection{Implementation and Pipeline}

Both architectures process the time series as follows:
\begin{itemize}[itemsep=0pt, parsep=0pt]
    \item \textbf{Preprocessing}: Remove apparent seasonality or fit basic model to compute residuals (if needed) \parencite{brockwell2016}.
    \item \textbf{Compute anomaly scores}: A scoring function (e.g., absolute changes, reconstruction errors from an autoencoder \parencite{ahmad2017}) is streamed.
    \item \textbf{Segmentation (SCS only)}: Segment incoming data by APCA or K-means.
    \item \textbf{Adaptive thresholding}:
    \begin{itemize}
        \item Update segment- or scale-specific confidence sequences.
        \item Optionally apply additional percentile or mixture model filtering \parencite{rousseeuw2020}.
    \end{itemize}
    \item \textbf{Decision layer}: Flag anomalies using composite rules.
\end{itemize}

\section{Experimental Results}

We evaluated both SCS and MACS against traditional and state-of-the-art adaptive methods on public datasets containing ground-truth anomaly labels. Metrics include the confusion matrix, change in accuracy, precision, recall, and F1-score compared to baseline. The experiments run from July 5th, 2025, to July 31st, 2025, over a month.

\subsection{Experiment and Dataset Description}

\subsubsection{Baseline: Traditional Percentile Thresholding}

Our reference method follows the classic \textit{p}-percentile rule.

\begin{enumerate}
    \item \textbf{Reconstruction-error vector}

    Let $x_t^{\prime}$ be the output of the diffusion auto-encoder at time $t$ and $x_t$ the original series window. We compute the point-wise L2 residual:
    \[
    r_t = \lVert x_t - x_t^{\prime} \rVert_2
    \]

    \item \textbf{Threshold selection}

    A global cut-off is chosen as the 99th percentile of the residual distribution on the training split:
    \[
    \theta = \texttt{Percentile}_{0.99}(\{r_t\}_{\text{train}})
    \]

    \item \textbf{Decision rule}

    A time stamp is labelled anomalous iff $r_t > \theta$.
\end{enumerate}

Although computationally trivial, this fixed‐quantile rule cannot adapt to regime shifts or changes in error variance – motivating the adaptive approaches studied in the remainder of the paper.

\subsubsection{Datasets}

\begin{table}[H]
\centering
\begin{tabular}{|p{3.0cm}|p{7.0cm}|p{4.2cm}|}
\hline
\textbf{Name of Dataset} & \textbf{Source \& Scope} & \textbf{Anomaly Labels} \\
\hline
Wafer Manufacturing & 151 inline process-control traces recorded by semiconductor sensors during wafer fabrication & Pass/fail ground truth from fab test lines ($\approx$10\% defective) \\
\hline
CalIt2 & People-count sensor at the main entrance of UC-Irvine’s CalIt2 building (15 weeks, 48 half-hour slots per day) & Event file with periods of abnormally high footfall (e.g., conferences) \\
\hline
Google Cloud Platform (GCP) & 30 service-category KPIs (service0–service29) collected from NVIDIA’s internal DGX-Cloud deployments & Manually curated incident tickets \\
\hline
Mars Science Laboratory (MSL) & NASA Mars Science Laboratory – 55 telemetry channels from Curiosity rover & 73729 test points with labelled off-nominal events (10.7\% anomalous) \\
\hline
Server Machine Dataset (SMD) & 5-week trace from 28 production servers, 38 KPIs each & Point-level labels (4.2\% anomaly) and attribution masks \\
\hline
CPU-KPI & Seasonal CPU-utilisation KPI released with Donut (public AIOps benchmark) & Partial point labels from capacity-planning alerts \\
\hline
\end{tabular}
\caption{Overview of evaluated datasets}
\end{table}

\begin{figure}[H]
    \centering
    \includegraphics[width=0.9\linewidth]{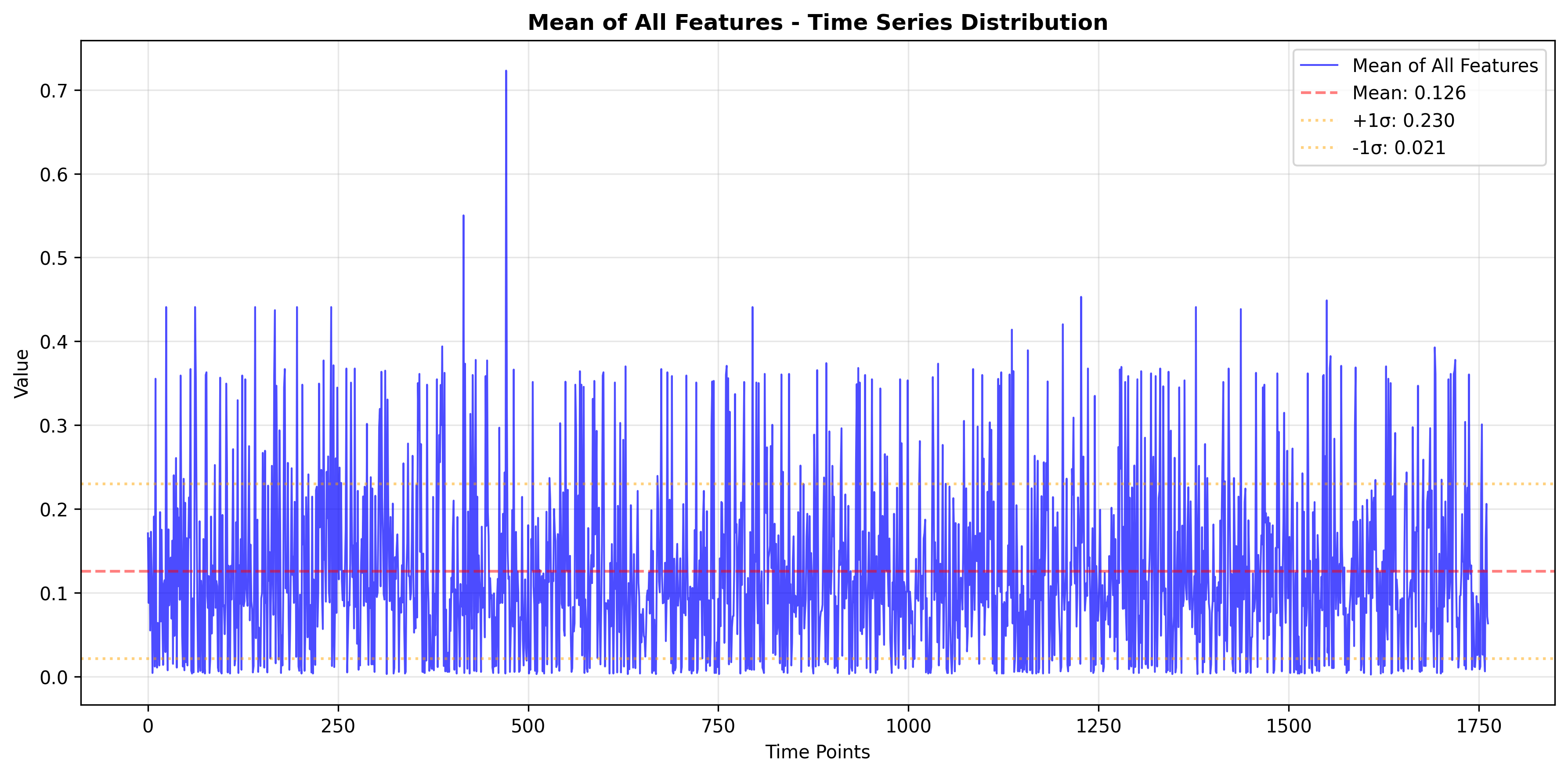}
    \caption{Illustration of Wafer Manufacturing dataset distribution}
    \label{fig:fig3}
\end{figure}

(Details for all dataset distribution are presented in \textit{Appendix D}.)

\subsubsection{Hyper-parameters and Variants}

\begin{itemize}[itemsep=0pt, parsep=0pt]
    \item Confidence level $1 - \alpha$ for adaptive confidence sequences: $\{0.05, 0.01\}$.
    \item Segmentation for SCS: Adaptive Piecewise Constant Approximation (APCA) vs. k-means on residual variance.
    \item Baseline: fixed 99\% percentile rule described in § 4.1.
\end{itemize}

\subsubsection{Evaluation Protocol}

For every dataset we compute:
\begin{itemize}[itemsep=0pt, parsep=0pt]
    \item Confusion-matrix counts (TP, FP, TN, FN)
    \item Change in Accuracy, Precision, Recall, F1 compared to baseline
    \item Proportional improvement over the baseline, calculated as: 
    \[
    \frac{\texttt{new\_method} - \texttt{traditional\_method}}{\texttt{traditional\_method}}
    \]
\end{itemize}

\subsection{Quantitative Comparison}

\textbf{Key results (Wafer Manufacturing dataset):}

\begin{table}[H]
\centering
\begin{tabular}{|l|c|c|c|c|}
\hline
\textbf{Method} & $\Delta$ Accuracy & $\Delta$ Precision & $\Delta$ Recall & $\Delta$ F1-Score \\
\hline
SCS APCA ($\alpha = 0.99$) & -0.0422 & -0.3282 & 3.9952 & 1.9074 \\
SCS KMEANS ($\alpha = 0.99$) & -0.0260 & -0.3999 & 1.6643 & 0.9262 \\
MACS Multi-Scale ($\alpha = 0.99$) & -0.0279 & -0.1890 & 3.9952 & 2.1705 \\
SCS APCA ($\alpha = 0.95$) & -0.0830 & -0.4290 & 6.1595 & 2.1289 \\
SCS KMEANS ($\alpha = 0.95$) & -0.0545 & -0.4656 & 3.3286 & 1.4148 \\
MACS Multi-Scale ($\alpha = 0.95$) & -0.0638 & -0.3651 & 5.6595 & 2.2349 \\
\hline
\end{tabular}
\caption{Performance delta on Wafer Manufacturing dataset}
\end{table}

\noindent\textbf{Comparison of the Number of Anomalies Detected}

\begin{table}[H]
\centering
\begin{tabular}{|l|c|c|c|c|}
\hline
\textbf{Method} & TP & TN & FP & FN \\
\hline
Traditional Percentile (99th percentile) & 6 & 1608 & 12 & 137 \\
SCS APCA ($\alpha = 0.99$) & 30 & 1516 & 104 & 113 \\
SCS KMEANS ($\alpha = 0.99$) & 16 & 1556 & 64 & 127 \\
MACS Multi-Scale ($\alpha = 0.99$) & 30 & 1539 & 81 & 113 \\
SCS APCA ($\alpha = 0.95$) & 43 & 1437 & 183 & 100 \\
SCS KMEANS ($\alpha = 0.95$) & 26 & 1500 & 120 & 117 \\
MACS Multi-Scale ($\alpha = 0.95$) & 40 & 1471 & 149 & 103 \\
\hline
\end{tabular}
\caption{Anomaly count comparison for Wafer Manufacturing dataset}
\end{table}

\begin{figure}[H]
    \centering
    \includegraphics[width=\linewidth]{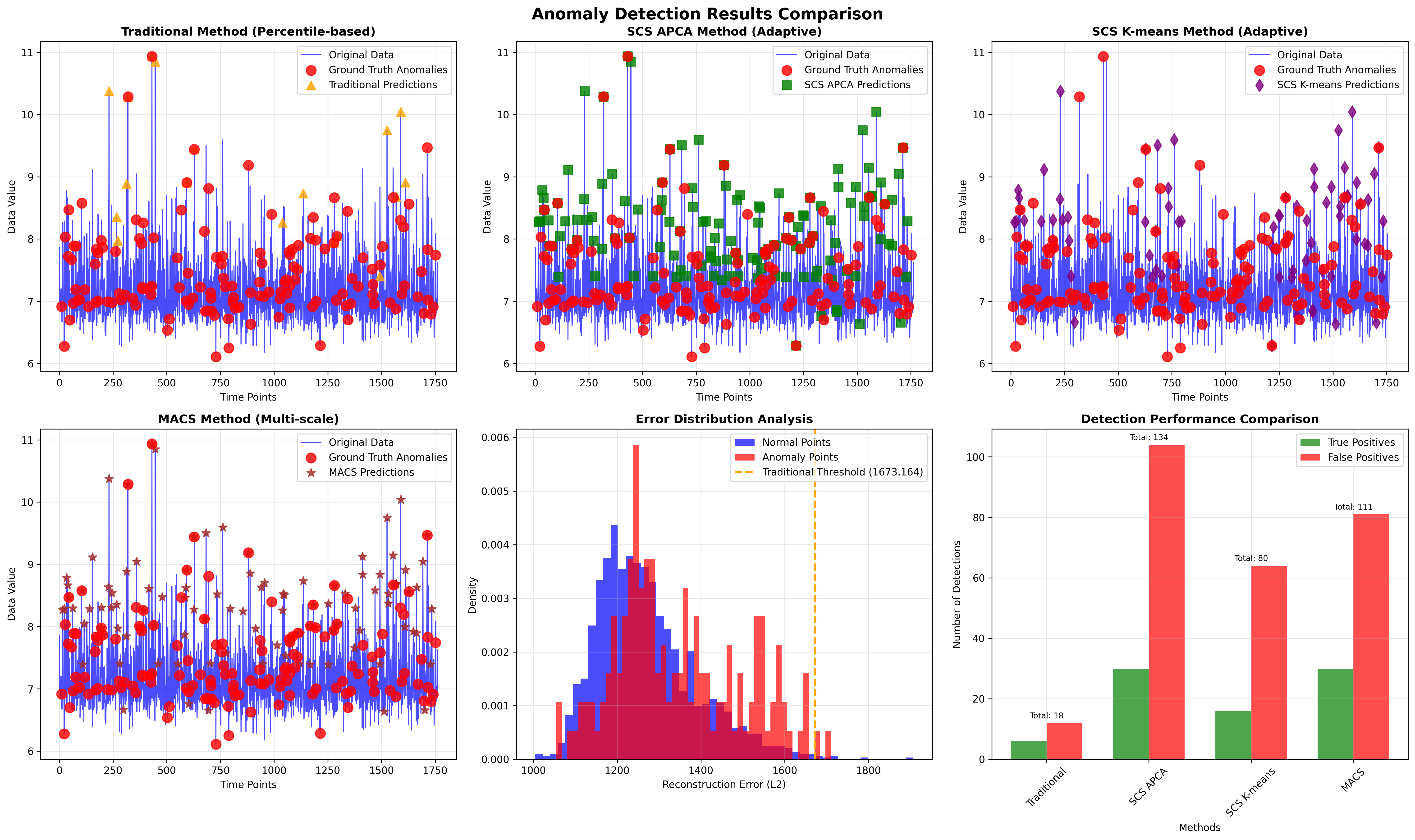}
    \caption{Illustration of results for Wafer Manufacturing dataset $\alpha = 0.99$}
    \label{fig:fig4}
\end{figure}

\begin{figure}[H]
    \centering
    \includegraphics[width=\linewidth]{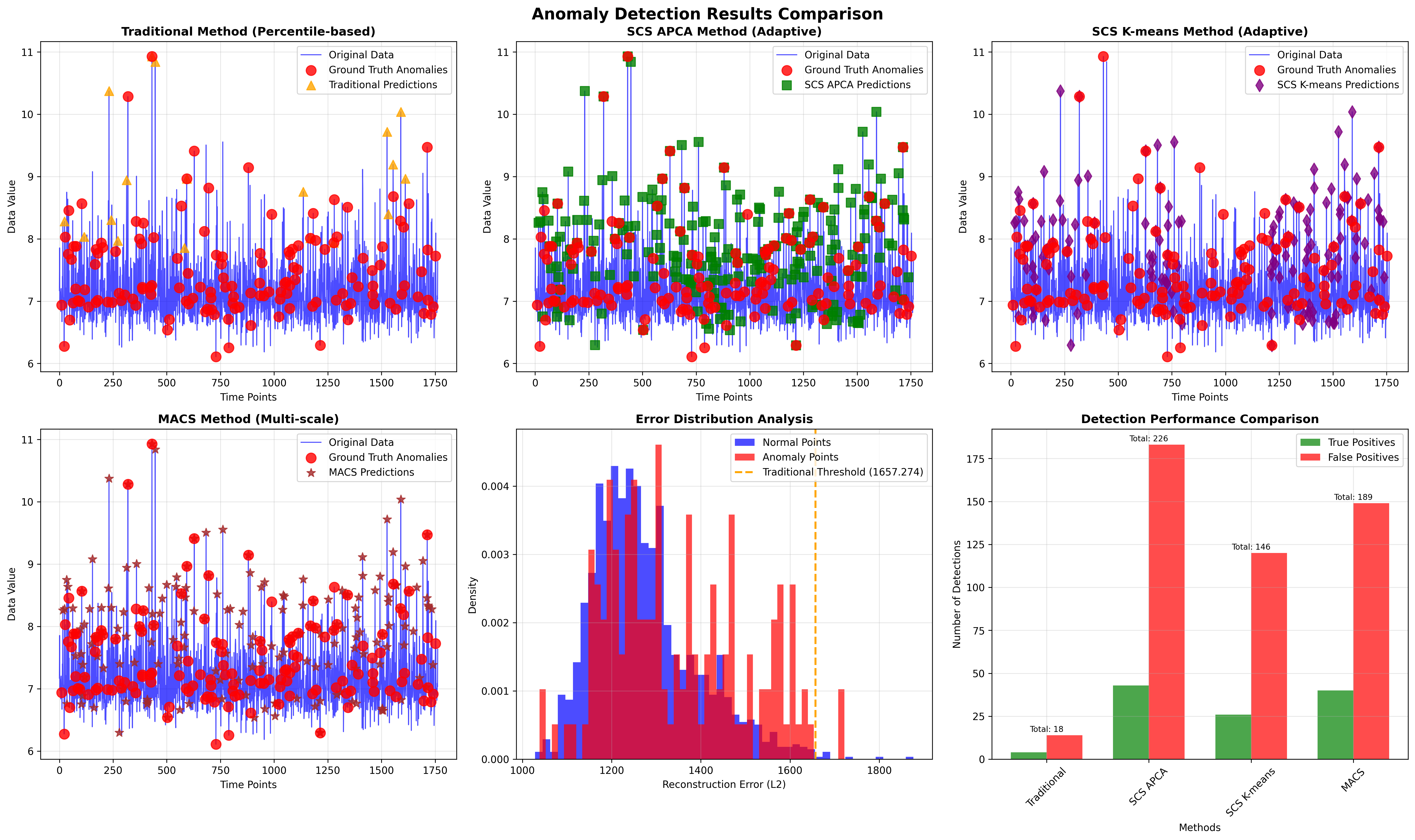}
    \caption{Illustration of results for Wafer Manufacturing dataset $\alpha = 0.95$}
    \label{fig:fig5}
\end{figure}

(Results on all public datasets are documented in the \textit{Appendix D}.)

\subsection{Detailed Analysis}

Both Segmented Confidence Sequences (SCS) and Multi-Scale Adaptive Confidence Segments (MACS) show significant performance improvements over the traditional static percentile thresholding approach across all evaluation metrics. Most notably, the F1-score of both SCS and MACS with a confidence level of $\alpha = 0.99$ increases approximately twice compared to the baseline, highlighting the benefit of adaptive, context-aware thresholds. When the confidence level is further reduced to $\alpha = 0.95$, recall improves substantially, leading to an over two times increase in F1-score relative to the baseline, even at the cost of a moderate decline in precision.

This trade-off between recall and precision reflects a typical pattern in adaptive detection: lowering the confidence threshold leads to more aggressive anomaly detection, capturing a larger proportion of true positives at the risk of including more false positives. Interestingly, this behavior is especially pronounced when the percentile filter is disabled. As shown in the anomaly count comparison, both SCS and MACS identify 30 true positive anomalies under $\alpha = 0.99$ with no filtering, which is a fivefold increase over the traditional method (which detects only six true positive anomalies).

In datasets with more complex temporal dynamics – such as sudden spikes, short bursts, or overlapping regimes – MACS is expected to outperform due to its ability to attend to fine-grained and coarse-grained deviations simultaneously. In contrast, SCS may be more effective when anomalies are aligned with persistent structural shifts, as it explicitly isolates and monitors regime-specific statistics.

The success of both approaches lies in their ability to localize statistical estimation. SCS adapts quickly to changes by segmenting the time series into regions with approximately stationary behavior, which allows for tight confidence bounds within each region. MACS, on the other hand, incorporates temporal diversity through rolling windows at multiple resolutions and adaptive attention weighting, enabling it to respond to anomalies that manifest at different time scales. Together, these techniques represent a principled move beyond static global thresholds and allow for a more interpretable, robust, and timely anomaly detection in real-world settings.

Finally, while removing the percentile filter maximizes recall and F1-score, this setting may not always be optimal in practice. In noisy environments or when false positives carry significant cost, reintroducing percentile filtering may be desirable to balance interpretability with operational reliability. Thus, both SCS and MACS offer flexible control over this trade-off depending on deployment constraints.

\section{Discussion}

Our empirical findings reinforce the known limitations of static thresholding techniques such as global percentiles and rolling quantiles when applied to nonstationary time series data. These traditional approaches fail to account for dynamic distributional shifts, leading to poor recall and under-detection of relevant anomalies \parencite{howard2021}. In contrast, the proposed SCS and MACS methods substantially improve performance by incorporating structural and temporal adaptivity. Specifically, they address evolving data behavior through segmentation (SCS) and multi-scale temporal analysis (MACS), yielding significant F1-score gains with only modest reductions in precision. 

\begin{figure}[H]
    \centering
    \includegraphics[width=0.6\linewidth]{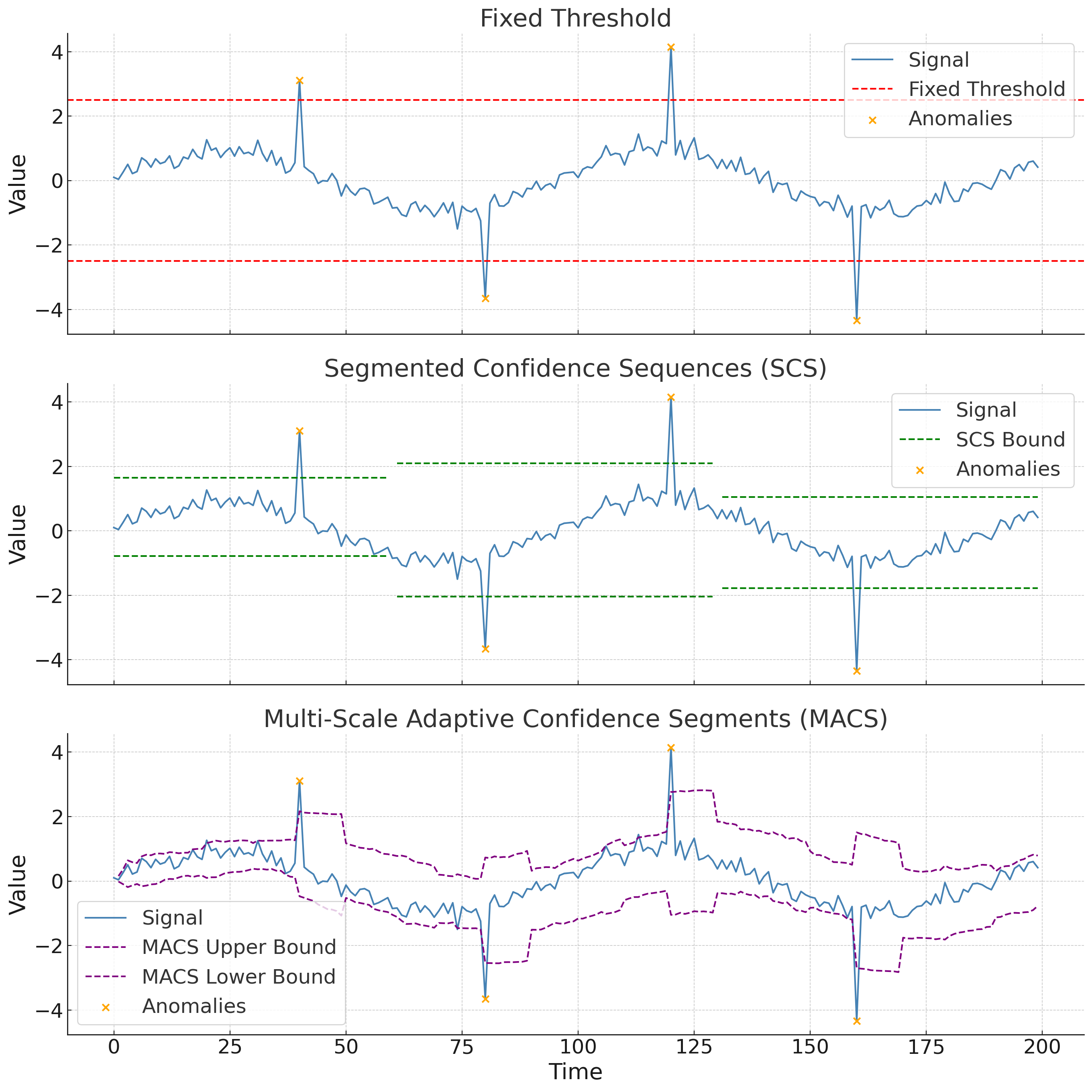}
    \caption{Illustration of different thresholding strategy}
    \label{fig:fig6}
\end{figure}

SCS is particularly well-suited to settings characterized by abrupt regime shifts and piecewise stationarity, where local adaptation via segmentation captures the changing statistical properties of the signal. Its regime-specific confidence sequences offer interpretable bounds and fast detection of contextual outliers. MACS, on the other hand, is more flexible across a wider range of temporal patterns. By leveraging multiple rolling windows and variance-sensitive attention mechanisms, MACS generalizes across both fast transients and slow drifts. This makes it especially effective in environments with layered or multi-scale anomaly behavior, such as bursty network activity or gradual process degradation \parencite{wang2023}.

A key advantage of both approaches lies in their model-free, unsupervised nature. Unlike many machine learning-based anomaly detectors, which often rely on labeled anomaly instances for training and hyperparameter tuning, SCS and MACS operate without supervision and retain explicit control over false alarm rates through statistically principled confidence sequences. This is crucial in high-stakes domains such as manufacturing, infrastructure monitoring, or cybersecurity, where excessive false positives can desensitize operators and degrade trust in automated systems \parencite{chandola2009, ahmad2017}.

Despite these advantages, our work also highlights some significant limitations and open challenges. The performance of SCS, in particular, is sensitive to the structure of the time series. In datasets that are highly stationary or exhibit noisy, unstructured behavior, segmentation may fail to produce meaningful partitions. Poorly defined segments can blur statistical distinctions and reduce detection quality. Similarly, while MACS benefits from its multi-scale architecture, its effectiveness hinges on the appropriate calibration of attention weights and confidence levels – parameters that may need tuning depending on the domain and noise profile. 

An important direction for future work is the development of robust online segmentation algorithms capable of operating under adversarial conditions or extreme nonstationarity. This includes detecting latent regime transitions that are subtle, overlapping, or induced by external interventions. Additionally, while this study used fixed window sizes for MACS, there is potential in exploring adaptive window scaling or learned attention mechanisms that adjust over time based on predictive uncertainty or performance feedback.

\section{Conclusion}

Adaptive thresholding is a critical component of reliable anomaly detection in nonstationary time series, where static baselines often fail to capture evolving data behavior. In this work, we introduced and systematically evaluated two novel frameworks – Segmented Confidence Sequences (SCS) and Multi-Scale Adaptive Confidence Segments (MACS) – that integrate online confidence sequence theory with localized statistical adaptation. By tailoring thresholding to the structure and scale of the data, both methods deliver statistically principled, interpretable, and high-performing anomaly detection. 

Our experimental results on benchmark Wafer Manufacturing datasets demonstrate that SCS and MACS significantly outperform traditional percentile-based and rolling quantile methods, particularly in terms of recall and F1-score. Both frameworks offer flexible precision-recall trade-offs through tunable confidence levels and percentile filtering, while maintaining robustness in unsupervised settings. 

Looking ahead, future work will explore extensions to multivariate time series, correlated or structured input streams, and integration with inference-based anomaly scoring methods. These directions aim to enhance further the expressiveness, generalizability, and deployment readiness of adaptive thresholding strategies for real-world anomaly detection.

\section{Acknowledgements}
We would like to thank Aaron Erickson, Saira Qureshi, Sanjeev Roka, Mahmoud Elhadidy, Sena Ekiz, and Jason Perlow for their insightful comments and guidance that significantly improved our work.

\printbibliography

\appendix
\section*{Appendix}

\subsection*{A. Pseudocode for Segmented Confidence Sequences (SCS)}

\begin{verbatim}
# Pseudocode for SCS adaptive thresholding

# Input: time_series, window_size, confidence_level, n_segments, segmentation_method

# Step 1: Segment the time series
if segmentation_method == "APCA":
    segments = APCA_segment(time_series, n_segments)
elif segmentation_method == "k-means":
    segments = kmeans_segment(time_series, n_segments)

# Step 2: Initialize confidence sequence per segment
for segment in segments:
    scores = compute_anomaly_scores(segment)
    conf_bounds = init_confidence_sequence(scores, confidence_level)

# Step 3: Online update and anomaly detection
for new_point in stream:
    assigned_segment = assign_to_segment(new_point, segments)
    update_confidence_sequence(assigned_segment, new_point)
    if is_anomalous(new_point, assigned_segment.conf_bounds):
        flag_anomaly(new_point)
\end{verbatim}

\subsection*{B. Pseudocode for Multi-Scale Adaptive Confidence Segments (MACS)}

\begin{verbatim}
# Pseudocode for MACS

# Input: time_series, short_window, medium_window, long_window, confidence_level

# Step 1: Maintain sliding windows at multiple scales
scales = [short_window, medium_window, long_window]
for scale in scales:
    window_scores[scale] = initialize_window(scale)
    conf_bounds[scale] = init_confidence_sequence(window_scores[scale], confidence_level)

# Step 2: Online anomaly detection
for new_point in stream:
    for scale in scales:
        window_scores[scale].add(new_point)
        update_confidence_sequence(window_scores[scale], confidence_level)
    # Composite decision rule: count scale violations
    violation_count = sum(is_anomalous(new_point, conf_bounds[scale]) for scale in scales)
    if violation_count >= threshold:  # e.g., 2 out of 3
        flag_anomaly(new_point)
\end{verbatim}

\subsection*{C. Pipeline Diagram (Suggested Structure)}

\begin{enumerate}
    \item Input: Time Series Data
    \item Preprocessing: Remove seasonality/trend if needed
    \item Segmentation Module:
    \begin{itemize}
        \item APCA or k-means segmentation (SCS)
        \item Multi-scale rolling windows (MACS)
    \end{itemize}
    \item Adaptive Thresholding:
    \begin{itemize}
        \item Segment-specific/confidence sequence update (SCS)
        \item Multi-scale online bounds (MACS)
    \end{itemize}
    \item Composite Detection Layer:
    \begin{itemize}
        \item Dual filtering: confidence violation and global percentile
        \item Anomaly decision based on a composite rule
    \end{itemize}
\end{enumerate}

\subsection*{D. Full Results Table}

\noindent\textbf{Wafer Manufacturing dataset distribution}

\begin{figure}[H]
    \centering
    \includegraphics[width=\linewidth]{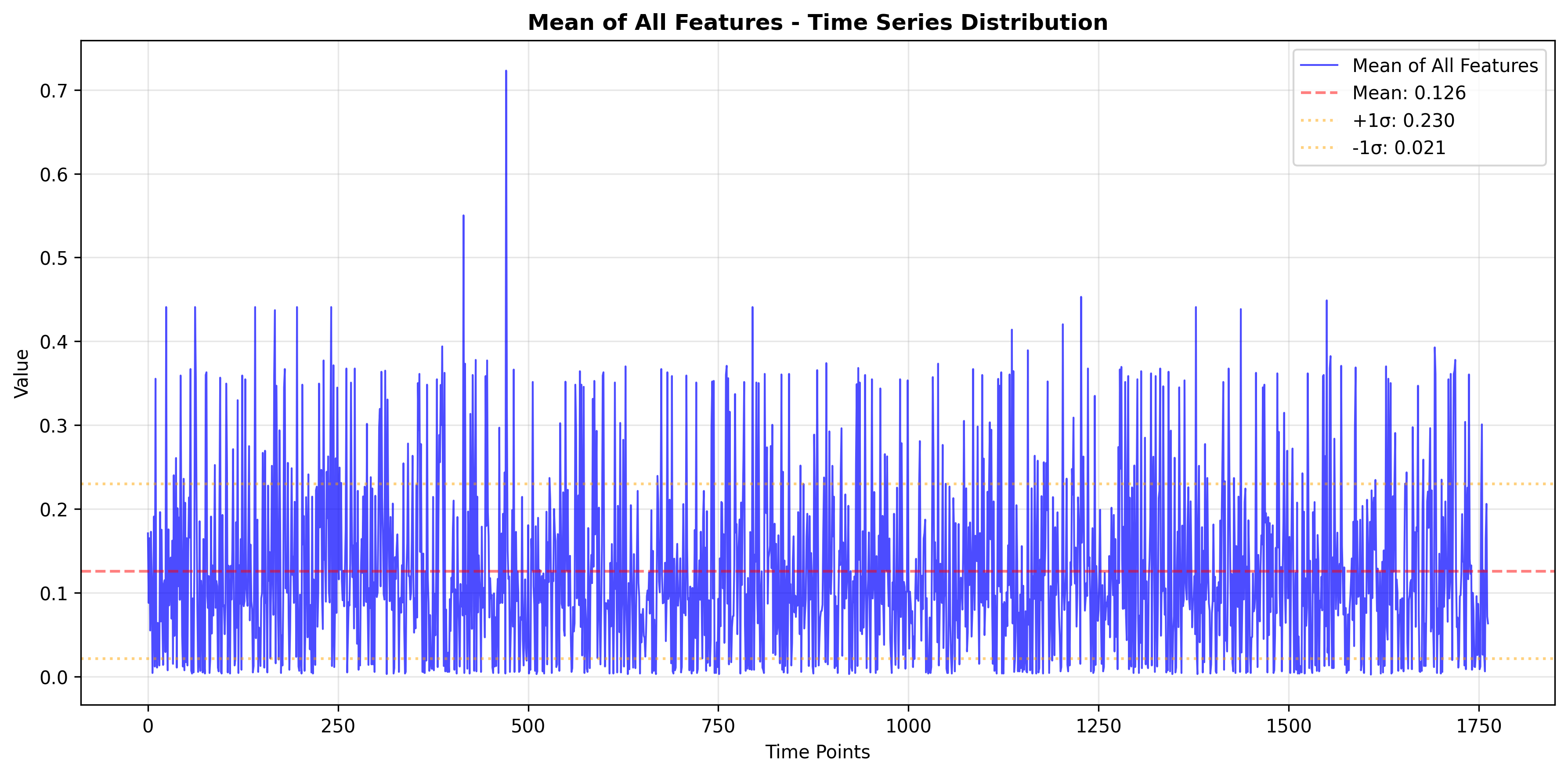}
\end{figure}

\noindent\textbf{Wafer Manufacturing dataset result}
\begin{table}[H]
\centering
\begin{tabular}{|l|r|r|r|r|}
\hline
\textbf{Method} & \textbf{$\Delta$ Accuracy} & \textbf{$\Delta$ Precision} & \textbf{$\Delta$ Recall} & \textbf{$\Delta$ F1-Score} \\
\hline
SCS APCA ($\alpha=0.99$)         & -0.0422 & -0.3282 & 3.9952 & 1.9074 \\
SCS KMEANS ($\alpha=0.99$)       & -0.0260 & -0.3999 & 1.6643 & 0.9262 \\
MACS Multi-Scale ($\alpha=0.99$) & -0.0279 & -0.1890 & 3.9952 & 2.1705 \\
SCS APCA ($\alpha=0.95$)         & -0.0830 & -0.4290 & 6.1595 & 2.1289 \\
SCS KMEANS ($\alpha=0.95$)       & -0.0545 & -0.4656 & 3.3286 & 1.4148 \\
MACS Multi-Scale ($\alpha=0.95$) & -0.0638 & -0.3651 & 5.6595 & 2.2349 \\
\hline
\end{tabular}
\end{table}

\noindent\textbf{Calit dataset distribution}

\begin{figure}[H]
    \centering
    \includegraphics[width=\linewidth]{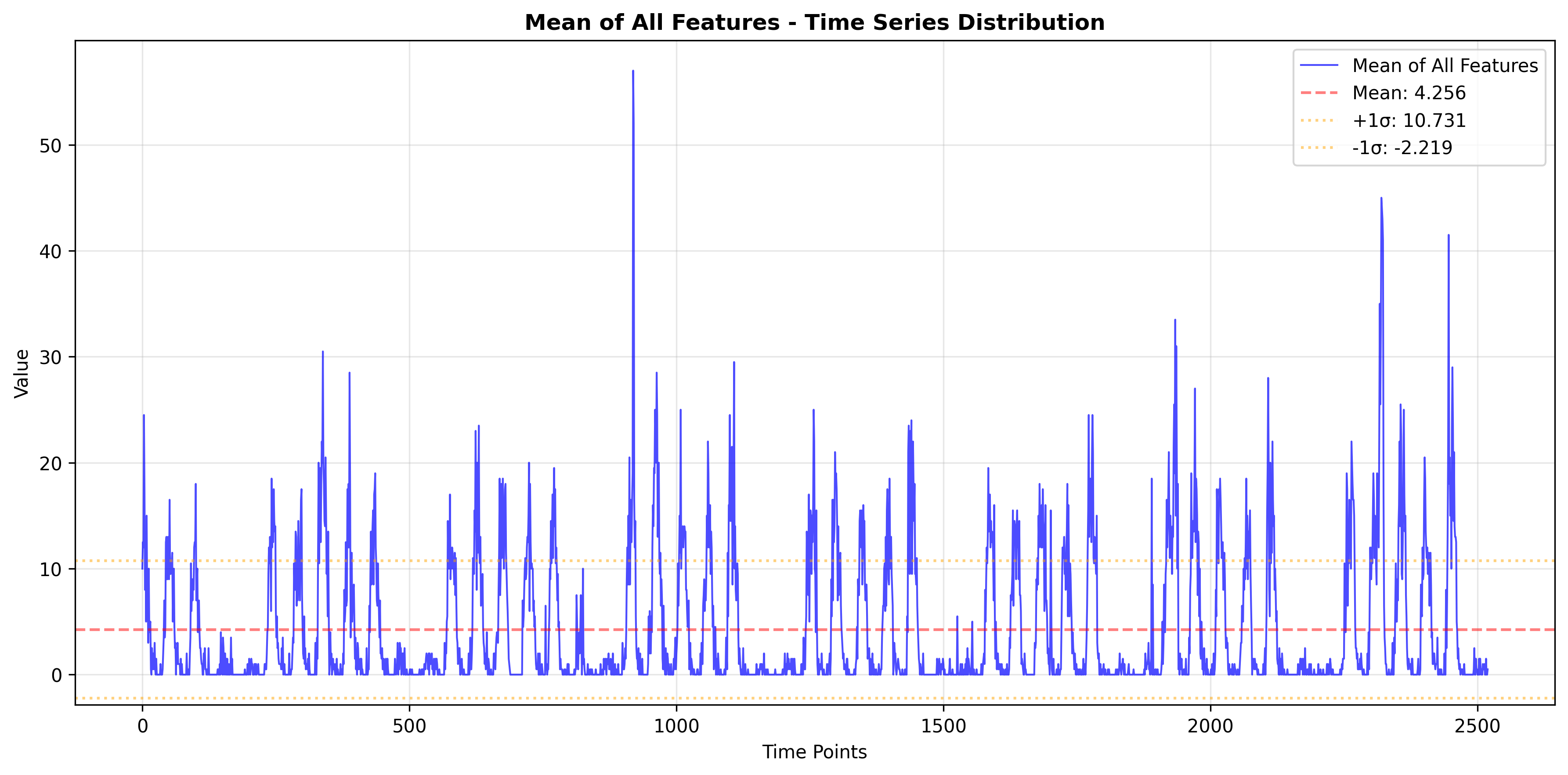}
\end{figure}

\noindent\textbf{Calit dataset result}
\begin{table}[H]
\centering
\begin{tabular}{|l|r|r|r|r|}
\hline
\textbf{Method} & \textbf{$\Delta$ Accuracy} & \textbf{$\Delta$ Precision} & \textbf{$\Delta$ Recall} & \textbf{$\Delta$ F1-Score} \\
\hline
SCS APCA ($\alpha=0.99$)         & -0.0542 & -0.4799 & 3.0000 & 0.4600 \\
SCS KMEANS ($\alpha=0.99$)       & -0.0518 & -0.3990 & 3.7135 & 0.6957 \\
MACS Multi-Scale ($\alpha=0.99$) & -0.0542 & -0.4799 & 3.0000 & 0.4600 \\
SCS APCA ($\alpha=0.95$)         & -0.0772 & -0.5286 & 3.8573 & 0.4200 \\
SCS KMEANS ($\alpha=0.95$)       & -0.0933 & -0.5981 & 3.7135 & 0.2436 \\
MACS Multi-Scale ($\alpha=0.95$) & -0.0772 & -0.5286 & 3.8573 & 0.4200 \\
\hline
\end{tabular}
\end{table}

\noindent\textbf{GCP dataset distribution}

\begin{figure}[H]
    \centering
    \includegraphics[width=\linewidth]{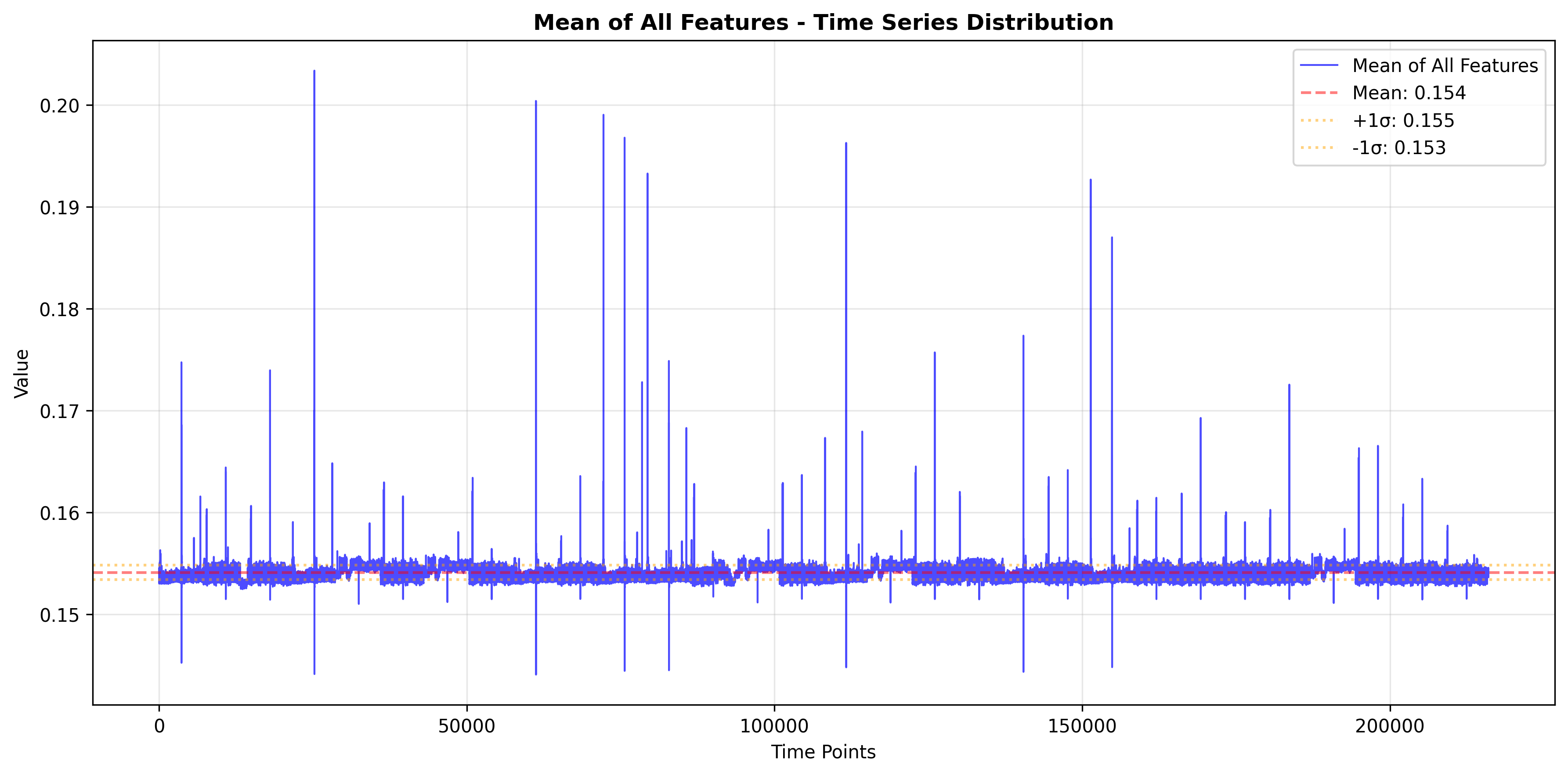}
\end{figure}

\noindent\textbf{GCP dataset result}
\begin{table}[H]
\centering
\begin{tabular}{|l|r|r|r|r|}
\hline
\textbf{Method} & \textbf{$\Delta$ Accuracy} & \textbf{$\Delta$ Precision} & \textbf{$\Delta$ Recall} & \textbf{$\Delta$ F1-Score} \\
\hline
SCS APCA ($\alpha=0.99$)         & -0.0463 &  0.0585 &  6.5054 & 4.8418 \\
SCS KMEANS ($\alpha=0.99$)       & -0.0073 &  0.2319 &  1.7527 & 1.6045 \\
MACS Multi-Scale ($\alpha=0.99$) & -0.0463 &  0.0585 &  6.5054 & 4.8418 \\
SCS APCA ($\alpha=0.95$)         & -0.0923 &  0.0654 & 13.0645 & 7.9435 \\
SCS KMEANS ($\alpha=0.95$)       & -0.0193 &  0.2723 &  4.2473 & 3.5819 \\
MACS Multi-Scale ($\alpha=0.95$) & -0.0923 &  0.0654 & 13.0645 & 7.9435 \\
\hline
\end{tabular}
\end{table}

\noindent\textbf{MSL dataset distribution}

\begin{figure}[H]
    \centering
    \includegraphics[width=\linewidth]{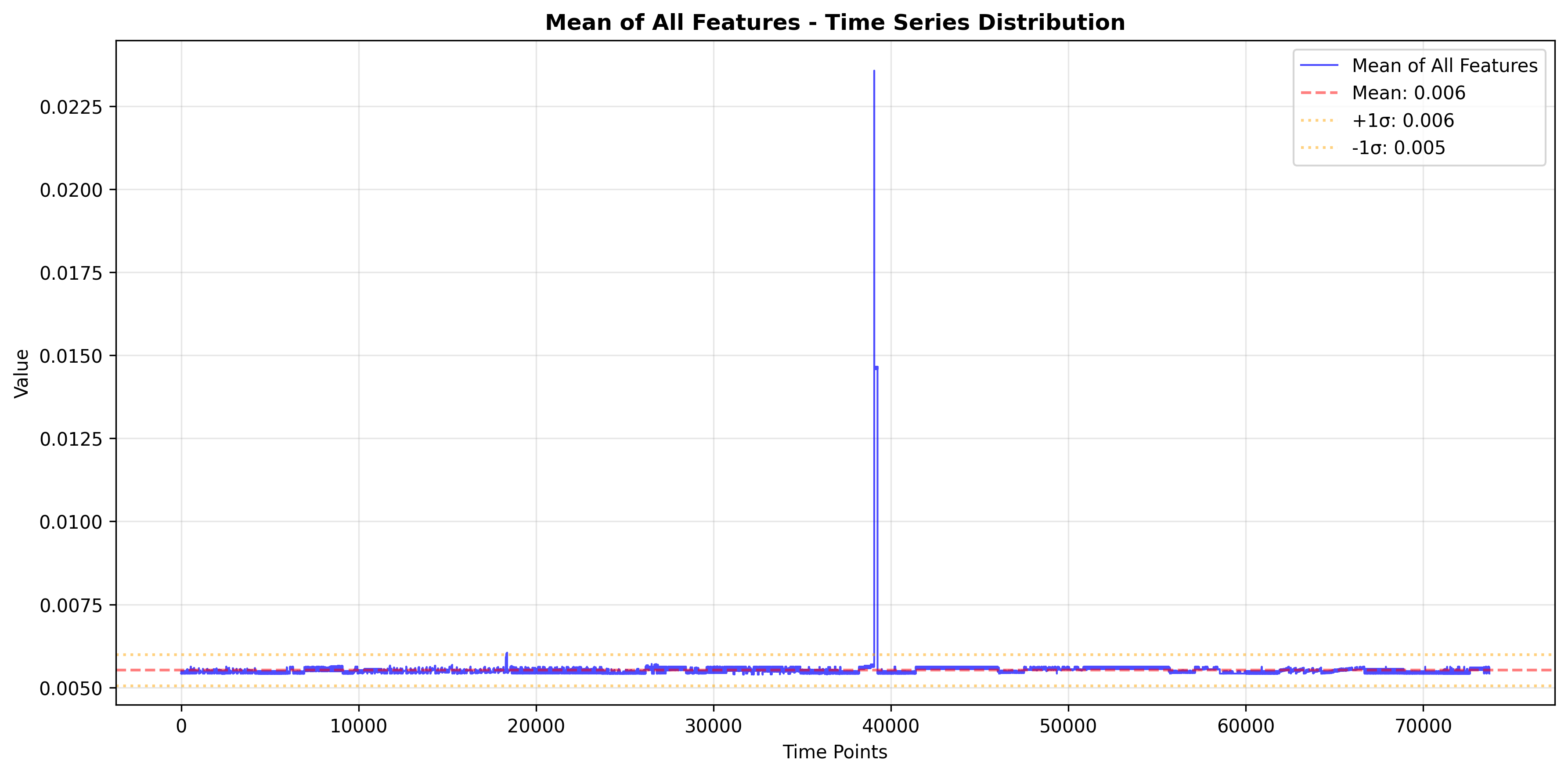}
\end{figure}

\noindent\textbf{MSL dataset result}

\begin{table}[H]
\centering
\begin{tabular}{|l|r|r|r|r|}
\hline
\textbf{Method} & \textbf{$\Delta$ Accuracy} & \textbf{$\Delta$ Precision} & \textbf{$\Delta$ Recall} & \textbf{$\Delta$ F1-Score} \\
\hline
SCS APCA ($\alpha=0.99$)         & -0.0710 & -0.0070 &  8.0741 & 4.2980 \\
SCS KMEANS ($\alpha=0.99$)       &  0.0005 &  0.2847 &  0.3333 & 0.3283 \\
MACS Multi-Scale ($\alpha=0.99$) & -0.0710 & -0.0070 &  8.0741 & 4.2980 \\
SCS APCA ($\alpha=0.95$)         & -0.1109 & -0.0685 & 11.5556 & 5.0101 \\
SCS KMEANS ($\alpha=0.95$)       & -0.0106 &  0.2276 &  1.9352 & 1.6061 \\
MACS Multi-Scale ($\alpha=0.95$) & -0.1084 & -0.0633 & 11.3611 & 4.9848 \\
\hline
\end{tabular}
\end{table}

\noindent\textbf{SMD dataset distribution}

\begin{figure}[H]
    \centering
    \includegraphics[width=\linewidth]{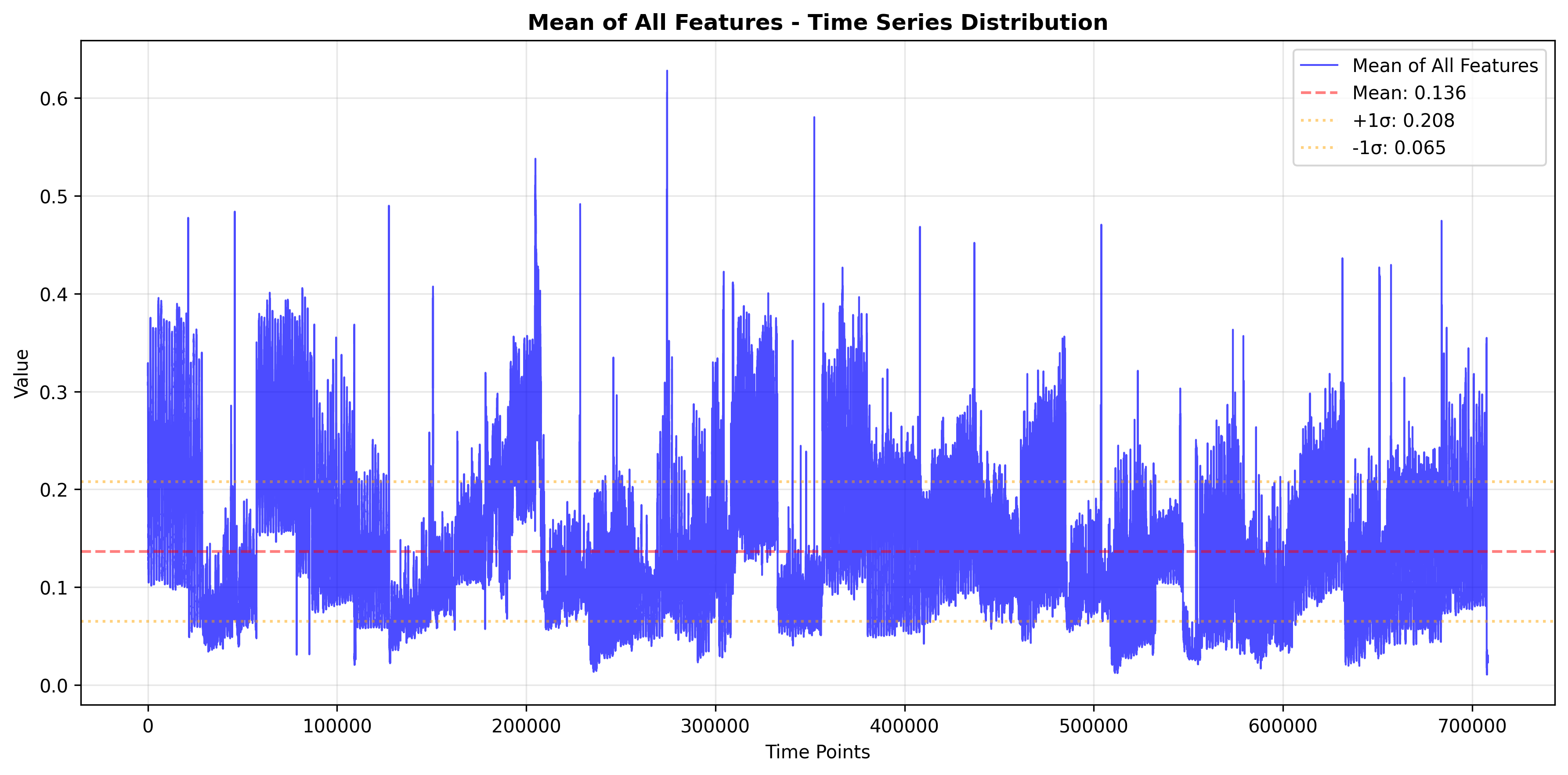}
\end{figure}

\noindent\textbf{SMD dataset result}

\begin{table}[H]
\centering
\begin{tabular}{|l|r|r|r|r|}
\hline
\textbf{Method} & \textbf{$\Delta$ Accuracy} & \textbf{$\Delta$ Precision} & \textbf{$\Delta$ Recall} & \textbf{$\Delta$ F1-Score} \\
\hline
SCS APCA ($\alpha=0.99$)         & -0.0594 &  0.4030 &  9.2152 & 3.5938 \\
SCS KMEANS ($\alpha=0.99$)       & -0.0146 &  0.4848 &  2.8481 & 1.9297 \\
MACS Multi-Scale ($\alpha=0.99$) & -0.0596 &  0.3576 &  8.8734 & 3.4453 \\
SCS APCA ($\alpha=0.95$)         & -0.1199 &  0.3758 & 17.7342 & 4.4297 \\
SCS KMEANS ($\alpha=0.95$)       & -0.0376 &  0.5091 &  6.5823 & 3.2500 \\
MACS Multi-Scale ($\alpha=0.95$) & -0.1198 &  0.3455 & 17.2785 & 4.3125 \\
\hline
\end{tabular}
\end{table}

\noindent\textbf{CPU dataset distribution}

\begin{figure}[H]
    \centering
    \includegraphics[width=\linewidth]{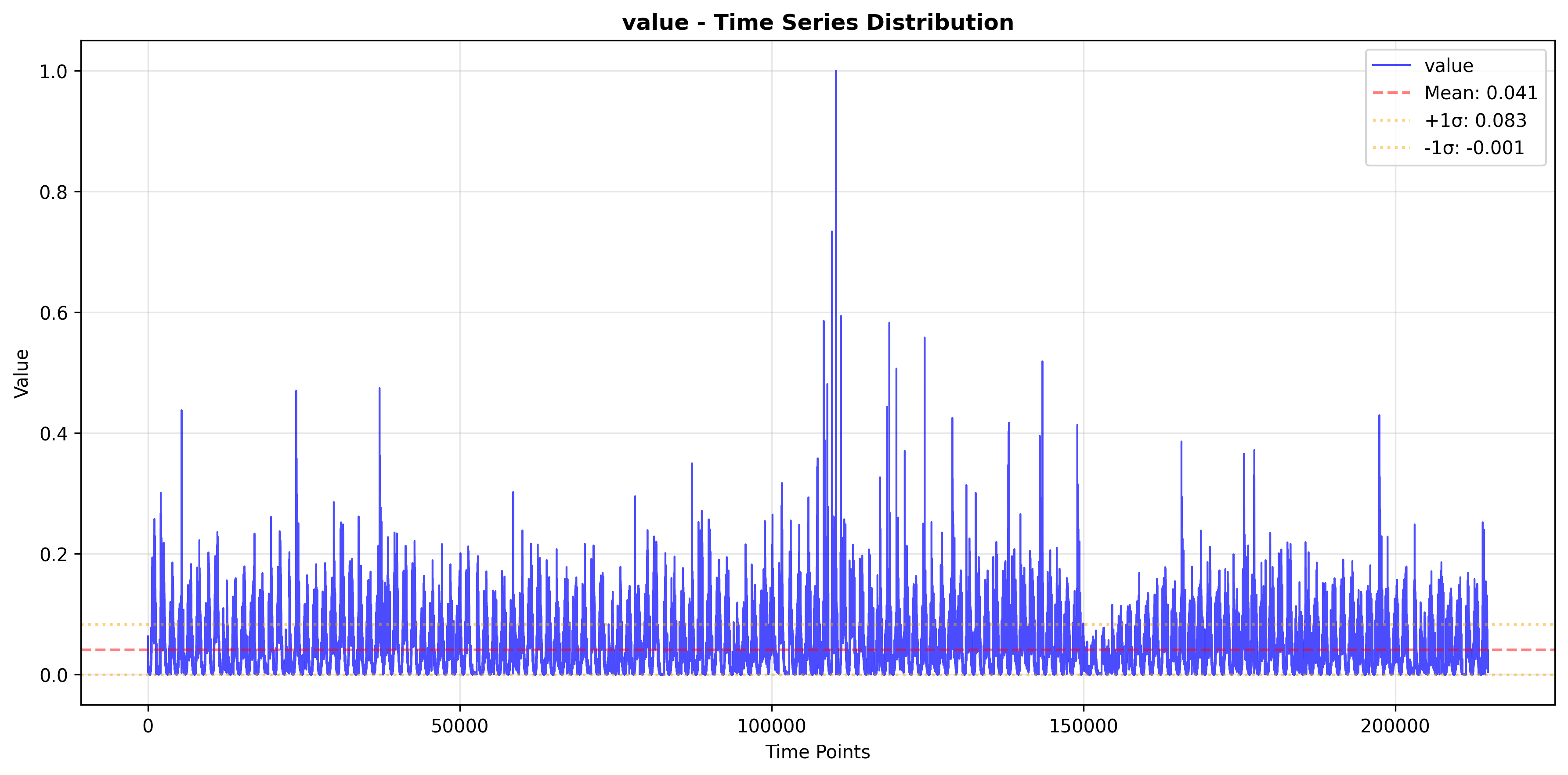}
\end{figure}

\noindent\textbf{CPU dataset result}

\begin{table}[H]
\centering
\begin{tabular}{|l|r|r|r|r|}
\hline
\textbf{Method} & \textbf{$\Delta$ Accuracy} & \textbf{$\Delta$ Precision} & \textbf{$\Delta$ Recall} & \textbf{$\Delta$ F1-Score} \\
\hline
SCS APCA ($\alpha=0.99$)         & -0.0538 & -0.4667 &  2.5971 & 1.1456 \\
SCS KMEANS ($\alpha=0.99$)       & -0.0110 & -0.5353 & -0.0777 & -0.1758 \\
MACS Multi-Scale ($\alpha=0.99$) & -0.0522 & -0.4796 &  2.3932 & 1.0549 \\
SCS APCA ($\alpha=0.95$)         & -0.1084 & -0.4867 &  5.4903 & 1.7335 \\
SCS KMEANS ($\alpha=0.95$)       & -0.0279 & -0.5495 &  0.7039 & 0.2802 \\
MACS Multi-Scale ($\alpha=0.95$) & -0.1064 & -0.4913 &  5.3155 & 1.6923 \\
\hline
\end{tabular}
\end{table}

\end{document}